%% file: MetaRuleGPT.tex
\documentclass[conference]{IEEEtran}
\IEEEoverridecommandlockouts
\usepackage{cite}
\usepackage{amsmath,amssymb,amsfonts}
\usepackage{algorithmic}
\usepackage{graphicx}
\usepackage{textcomp}
\usepackage{xcolor}

\usepackage{hyperref} % 引用会有高亮
\newcommand{\ff}[1]{$ #1$}   % math environment + newline

\newcommand{\bri}[1]{\uppercase\expandafter{\romannumeral#1}}

\newcommand{\everything}[1]{\begin{document}\begin{CJK*}{UTF8}{gkai}#1\end{CJK*}\end{document}}

\newcommand{\cccode}[1]{\begin{lstlisting}#1\end{lstlisting}}

\newcommand{\ttt}{\subsection{}}

\newcommand{\eee}[1]{\begin{enumerate}#1\end{enumerate}}

\newcommand{\pdfFrac}[2]{\frac{\partial #1}{\partial #2}}
\newcommand{\OFL}{\mathrm{OFL}}
\newcommand{\UFL}{\mathrm{UFL}}
\newcommand{\fl}{\mathrm{fl}}
\newcommand{\op}{\odot}
\newcommand{\Eabs}{E_{\mathrm{abs}}}
\newcommand{\Erel}{E_{\mathrm{rel}}}
\newcommand{\red}{\textcolor{red}}
\newcommand{\blue}{\textcolor{blue}}
\newcommand{\hl}[1]{\textcolor{red}{#1}}
\def\BibTeX{{\rm B\kern-.05em{\sc i\kern-.025em b}\kern-.08em
    T\kern-.1667em\lower.7ex\hbox{E}\kern-.125emX}}
\begin{document}

\title{MetaRuleGPT: Recursive Numerical Reasoning of Language Models Trained with Simple Rules
\\

\thanks{* Equal Contribution}
\thanks{$\dag$ Corresponding Author}
}
%\title{Enhancing Numerical Reasoning in Language Models through Rule Learning}

\author{\IEEEauthorblockN{Kejie Chen\textsuperscript{*}}
\IEEEauthorblockA{\textit{Zhejiang University}\\
Hangzhou, China \\
22135030@zju.edu.cn}
\and
\IEEEauthorblockN{Lin Wang\textsuperscript{*}}
\IEEEauthorblockA{
\textit{Zhejiang University}\\
Hangzhou, China \\
wanglin1126@zju.edu.cn}
\and
\IEEEauthorblockN{Qinghai Zhang\textsuperscript{$\dag$}}
\IEEEauthorblockA{
\textit{Zhejiang University} \\
Hangzhou, China \\
qinghai@zju.edu.cn}
\and
\IEEEauthorblockN{Renjun Xu\textsuperscript{$\dag$}}
\IEEEauthorblockA{\textit{Zhejiang University} \\
Hangzhou, China \\
rux@zju.edu.cn}
}

\maketitle
% \footnotetext[1]{† Corresponding author.}
\begin{abstract}
\input{contents/abstract}
\end{abstract}


\section{Introduction}
\input{contents/introduction}

\section{RELATED WORK}
\input{contents/related_work}

\section{Research Methodology}

\input{contents/research_methodology}
\section{EXPERIMENTS}
\input{contents/experiment}

\section{RESULTS AND DISCUSSION}
\input{contents/Results_and_Discussion}

\section{Conclusion}
This study explores the rule-following capabilities of language models, focusing on the combinatorial skills and generalization abilities that humans display in problem-solving. We introduce MetaRuleGPT, a Transformer-based language model that utilizes an iterative strategy and learns from a series of compound rules and sub-rules. With only 30 million parameters, MetaRuleGPT demonstrates high accuracy in handling high-digit calculations and complex vector cross-product operations, surpassing current mainstream large language models. Our findings highlight the importance of incorporating rule-based learning in language models to enhance their numerical reasoning abilities and generalization skills. MetaRuleGPT's success in solving complex mathematical problems with relatively few parameters showcases the effectiveness of this approach and paves the way for future research in this direction.

\section*{Limitations} 
This research raises several issues worthy of further exploration, including:
\eee{
\item Although our model has shown certain generalization and understanding abilities after rule learning, it is limited by computational resources, and the variety of problems it can handle is relatively limited. Expanding the model's parameter size and training with more diverse rule datasets could enable MetaRuleGPT to handle a broader range of logical tasks.

\item The controllability of MetaRuleGPT may vary with the increase in required tasks, and we plan to add more task rules in future model training to further evaluate and improve the model's controllability. Moreover, we aim to enhance MetaRuleGPT's controllability when handling tasks beyond its current capabilities. For instance, while MetaRuleGPT excels in numerical computation tasks, it may produce unpredictable results when attempting function integration problems, leading to significant errors. Addressing this challenge is one of our current focuses, and we plan to introduce more rule data in future optimizations to improve the model's overall controllability, enabling greater stability and accuracy across a wider range of applications.

%\item While MetaRuleGPT can generally mimic human problem-solving approaches for numerical issues, it still lacks human-like precision in certain details. For example, the model often treats zero mechanically, unable to recognize its special properties as humans do. This indicates a gap between the model's fine-grained problem-solving and human capabilities, which we aim to address through optimized learning strategies and refined issue handling.

\item MetaRuleGPT currently cannot automatically handle untrained generalization forms or novel concepts beyond the meta-learning distribution, which limits its ability to tackle entirely new problems. Achieving human-like systematic generalization by leveraging real-world training experiences remains an open question and a direction for future research.
}

% \section*{Acknowledgment}

% The preferred spelling of the word ``acknowledgment'' in America is without 
% an ``e'' after the ``g''. Avoid the stilted expression ``one of us (R. B. 
% G.) thanks $\ldots$''. Instead, try ``R. B. G. thanks$\ldots$''. Put sponsor 
% acknowledgments in the unnumbered footnote on the first page.

% \section*{References}

% Please number citations consecutively within brackets \cite{b1}. The 
% sentence punctuation follows the bracket \cite{b2}. Refer simply to the reference 
% number, as in \cite{b3}---do not use ``Ref. \cite{b3}'' or ``reference \cite{b3}'' except at 
% the beginning of a sentence: ``Reference \cite{b3} was the first $\ldots$''

% Number footnotes separately in superscripts. Place the actual footnote at 
% the bottom of the column in which it was cited. Do not put footnotes in the 
% abstract or reference list. Use letters for table footnotes.

% Unless there are six authors or more give all authors' names; do not use 
% ``et al.''. Papers that have not been published, even if they have been 
% submitted for publication, should be cited as ``unpublished'' \cite{b4}. Papers 
% that have been accepted for publication should be cited as ``in press'' \cite{b5}. 
% Capitalize only the first word in a paper title, except for proper nouns and 
% element symbols.

% For papers published in translation journals, please give the English 
% citation first, followed by the original foreign-language citation \cite{b6}.
\bibliographystyle{IEEEtran}
\bibliography{ref}

\end{document}

%% file: contents/abstract.tex
Recent studies have highlighted the limitations of large language models in mathematical reasoning, particularly their inability to capture the underlying logic. Inspired by meta-learning, we propose that models should acquire not only task-specific knowledge but also transferable problem-solving skills. We introduce MetaRuleGPT, a novel Transformer-based architecture that performs precise numerical calculations and complex logical operations by learning and combining different rules. In contrast with traditional training sets, which are heavily composed of massive raw instance data, MetaRuleGPT is pre-trained on much less abstract datasets containing basic, compound, and iterative rules for mathematical reasoning. Extensive experimental results demonstrate MetaRuleGPT can mimic human's rule-following capabilities, break down complexity, and iteratively derive accurate results for complex mathematical problems. These findings prove the potential of rule learning to enhance the numerical reasoning abilities of language models.

%Our experimental results demonstrate that MetaRuleGPT is superior in terms of high-digit calculations and vector cross products over state-of-the-art models. Moreover, 

%% file: contents/introduction.tex
In the field of Natural Language Processing, large language models such as GPT-4 \cite{lim2023benchmarking} have made remarkable progress, demonstrating impressive understanding and generation capabilities across a wide range of tasks such as text summarization\cite{volske-etal-2017-tl}\cite{hermann2015teachingmachinesreadcomprehend}\cite{narayan2018dontdetailsjustsummary} question answering\cite{bisk2019piqareasoningphysicalcommonsense}\cite{hendrycks2021measuringmassivemultitasklanguage} and mathematical reasoning\cite{yu2024metamathbootstrapmathematicalquestions}\cite{luo2023wizardmathempoweringmathematicalreasoning}\cite{wang2024mathshepherdverifyreinforcellms}. However, these models still face considerable challenges when encountering 
%mathematical problems and other domains requiring specialized knowledge. Mathematical tasks span 
mathematical tasks in a broad spectrum, including but not limited to basic arithmetic operations, calculus, and equation solving. Despite their powerful language understanding capabilities, these models often struggle with even simple numerical calculations. For example, given a straightforward addition problem:
$$      241257284+758742716=1000000000, $$
most current mainstream language models have difficulty in correctly completing such basic mathematical operations.

Inspired by the success of chain-of-thought(CoT)\cite{wei2023chainofthoughtpromptingelicitsreasoning} reasoning, we can see that, similar to the human brain's System II, a model can achieve more reliable results on complex input problems through repeated and recursive thinking. In mathematics, this approach—recursive reasoning with a simple set of rules—serves as an effective paradigm for precisely solving complex issues.

\begin{figure*}[tp]
  \centering 
  \includegraphics[width=\textwidth]{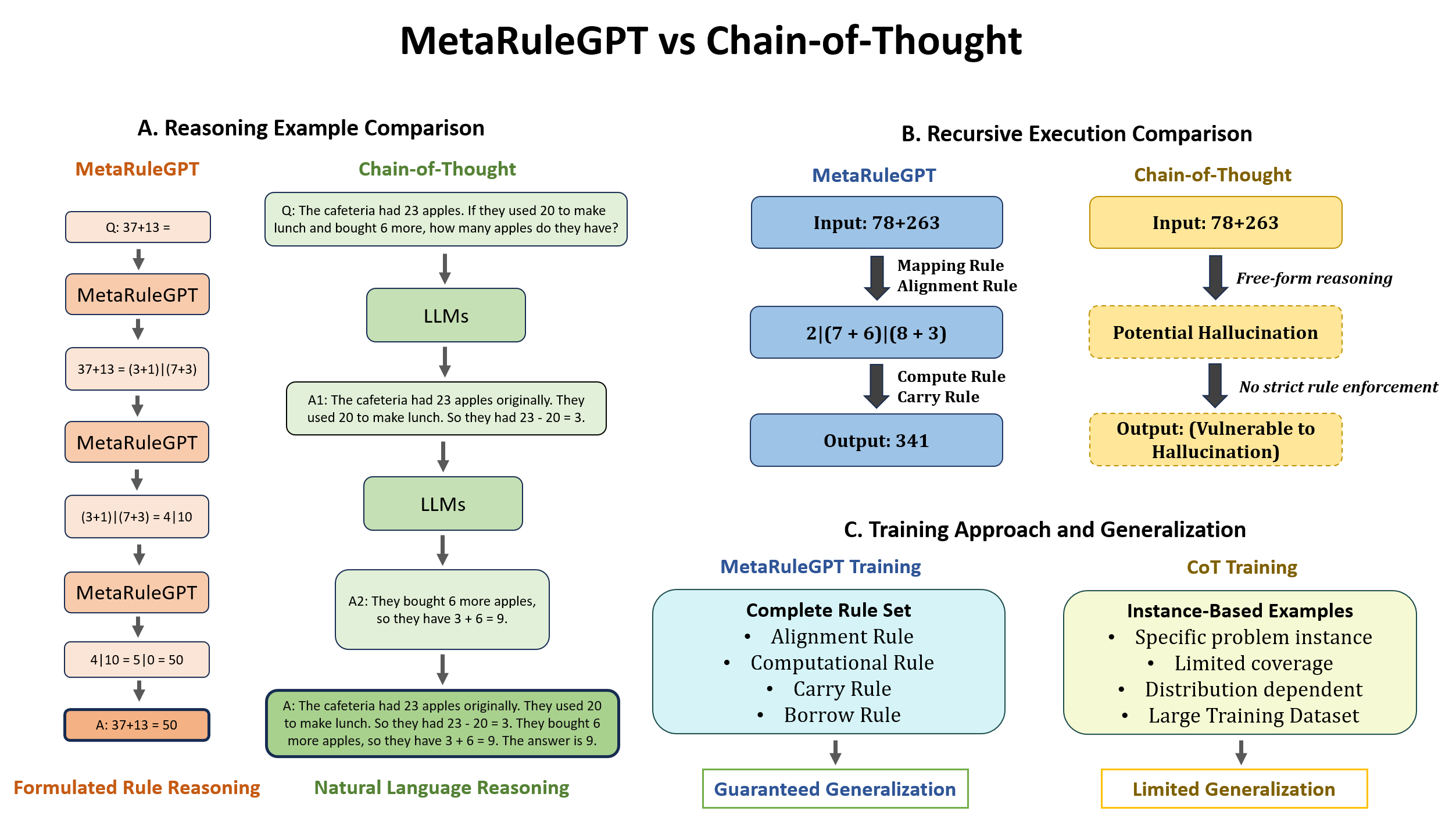}
  \caption{This image illustrates the differences between MetaRuleGPT and the traditional Chain-of-Thought (CoT) reasoning method in handling mathematical problems. MetaRuleGPT employs a rule-based reasoning approach, breaking down and solving problems step-by-step through predefined rules, such as alignment rules, carry rules, and borrow rules. This structured method ensures accuracy and generalization ability in reasoning, allowing the model to systematically handle various computational tasks while avoiding the common hallucination errors found in CoT methods.} 

  \label{fig:vs} 
\end{figure*}

In this context, we introduce MetaRuleGPT, a cutting-edge Transformer-based\cite{vaswani2017attention} architecture that excels CoT in precise numerical calculations and intricate logical operations through the learning and integration of diverse rules as shown in Fig~\ref{fig:vs}. MetaRuleGPT is adept at emulating human-like rule-following behaviors, enabling it to simplify complexities and iteratively arrive at accurate solutions for challenging mathematical problems. This advancement underscores the promise of rule learning in enhancing the numerical reasoning capabilities of language models.

%% file: contents/related_work.tex
A body of research indicates that LLMs often fail to capture the underlying logic required for mathematical problem-solving, particularly in tasks\cite{hendrycks2021measuringmassivemultitasklanguage}\cite{cobbe2021trainingverifierssolvemath} that necessitate precise numerical calculations and complex logical operations. Brown et al. \cite{brown2020language}, have examined the limitations of LLMs in mathematical tasks, revealing that even basic operations can lead to inaccuracies. A few methods\cite{zhou2023leasttomostpromptingenablescomplex}\cite{fu2023complexitybasedpromptingmultistepreasoning}\cite{wang2023selfconsistencyimproveschainthought} are proposed to improve the quality of reasoning paths. Complexity-based CoT\cite{fu2023complexitybasedpromptingmultistepreasoning} selects examples with more steps as in-context demonstrations and shows that prompting with more reasoning steps leads to better performance. Wei et al\cite{wei2022finetuned} explored the role of fine-tuning and prompt engineering in improving LLM performance on specific tasks, yet these methods often fall short when applied to higher-order mathematical reasoning.

Meta-learning\cite{vanschoren2018metalearning} has emerged as a promising approach to enhance model capabilities, emphasizing the need for transferable problem-solving skills rather than mere task-specific knowledge. Research by Finn et al. \cite{finn2017model} and subsequent meta-learning frameworks illustrate how models can be designed to generalize across tasks by learning from previous experiences. This paradigm has inspired our development of MetaRuleGPT, which not only seeks to improve numerical reasoning but also aims to equip models with the ability to learn and apply a variety of mathematical rules through structured pre-training.

Rule learning\cite{bassel2011rulelearning} demonstrated its efficacy in enhancing reasoning abilities, showing that integrating rule-based reasoning into model architectures can yield significant performance gains, especially in complex logical reasoning tasks. Building on these insights, MetaRuleGPT employs a Transformer-based architecture that systematically incorporates basic, compound, and iterative rules for mathematical reasoning, thereby addressing the shortcomings observed in existing models.

%% file: contents/research_methodology.tex
To enhance the accuracy and generalization of language models \cite{zhang2019ernie} in solving complex logical reasoning and numerical calculation tasks, we introduce the MetaRuleGPT model. This model aims to bolster the reasoning capabilities and generalization potential of language models \cite{lake2023human}, inspired by the concept of meta-learning. MetaRuleGPT focuses on mastering general learning strategies to precisely complete complex logical deduction tasks by applying learned rules. The model dynamically integrates basic mathematical computation rules with higher-order operation rules, enhancing its ability to process rule combinations. This design allows MetaRuleGPT to exhibit superior accuracy and generalization when faced with complex logical reasoning challenges, such as mathematical problems.

Furthermore, by adopting this strategy, MetaRuleGPT is not limited to handling a single task but is capable of learning and executing various different tasks. When dealing with multitasking, the rules across different tasks might intersect; our model can flexibly learn these intersecting rules and dynamically apply them to complete multiple tasks simultaneously, while keeping the tasks independent of each other without interference.

\subsection{Specific Rule Learning for Arithmetic Tasks}

\begin{table*}[h!] % 放置表格的位置，h表示here
  \centering % 居中
  %\small
  %% \caption{各任务学习规则汇总表} % 表格标题              % 
  \caption{Summary Table of Learning Rules for Various Tasks} % 表格标题 
  \begin{tabular}{ccccccc} % 三列，居中对齐，竖线分隔列
    \hline % 表格横线
    Rule Type & Numerical Addition & Numerical Subtraction & Vector Cross Product \\
    \hline  
    Vector Table    &  -  &  -  &  \ff{\surd } \\  
    Nine Addition Table    &  \ff{\surd }  &  -  &  \ff{\surd } \\
    Nine Subtraction Table   &  -  &  \ff{\surd }  &  \ff{\surd } \\
    Nine Multiplication Table &  -  &  -  &  \ff{\surd } \\
    Mapping Rule  &  \ff{\surd }  &  \ff{\surd }  &  \ff{\surd }\\
    Carrying Rule &  \ff{\surd }  &  -  &  \ff{\surd } \\
    Borrowing Rule &  -  &  \ff{\surd }  &  \ff{\surd } \\
    Vector Product Rule  &  -  &  -  &  \ff{\surd } \\
    Compute Rule  &  \ff{\surd }  &  \ff{\surd }  &  \ff{\surd } \\
    \hline
    \hline
  \end{tabular}
\label{tab:all_rule}
\end{table*}

% In our research, the model demonstrated outstanding logical reasoning and generalization capabilities in performing three complex tasks: high-digit(10 or more digits) addition and subtraction calculations, and vector cross-product computations. For these tasks, we designed specific rule datasets for training. For example, during the training for addition calculations, the model was guided to learn key knowledge including single-digit addition rules, carry rules, digit mapping rules, and basic computation rules. By mastering these basic rules, after meticulous pre-training \cite{zhang2019ernie}, our model could flexibly apply and combine these rules to accurately complete complex mathematical operations, including high-digit addition and subtraction.

\begin{figure}[tp]
  \centering
  \includegraphics[scale=0.35]{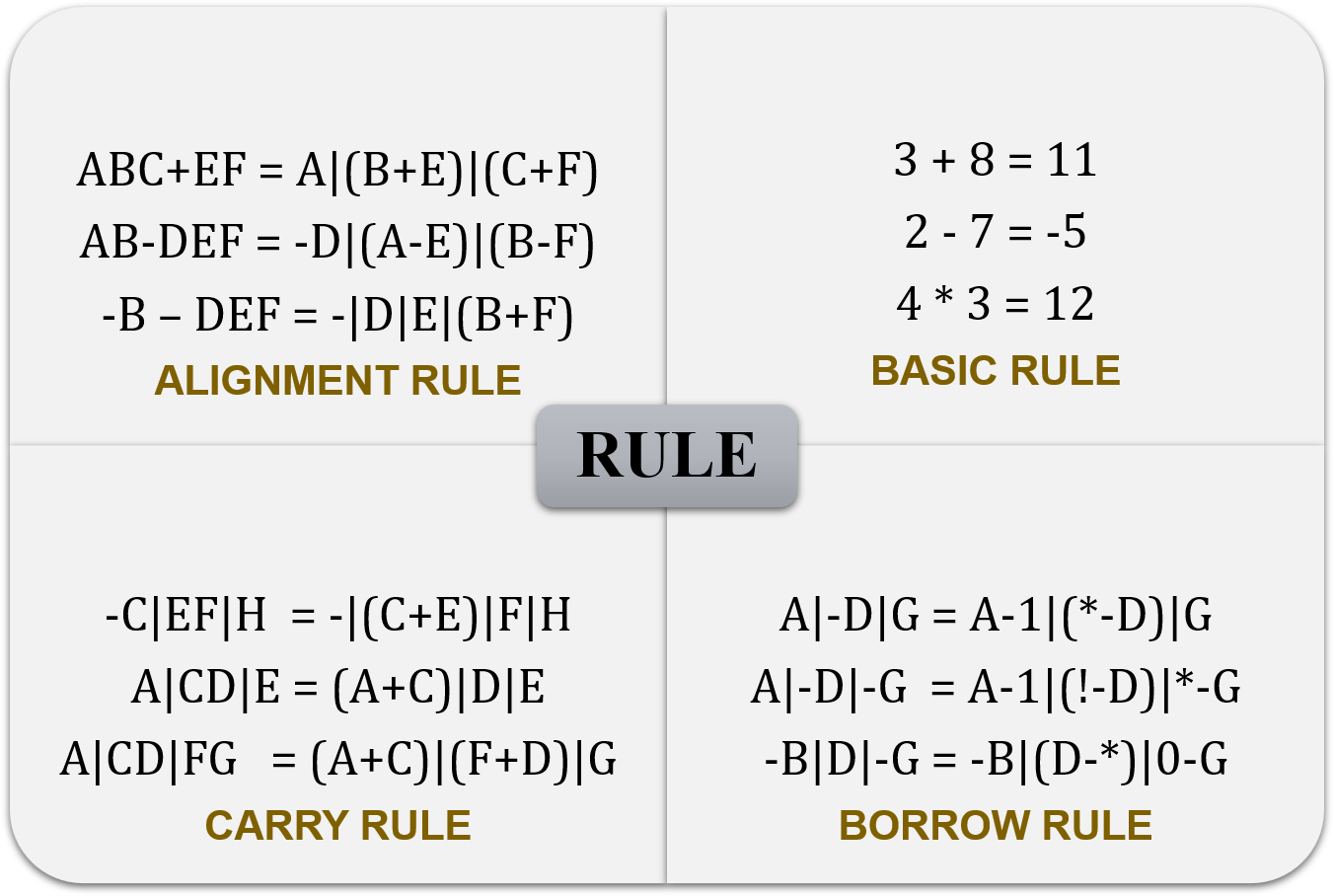}
  \caption{This picture shows some training rule examples, where * and ! represents 10 and 9 in decimal respectively. Once our training data covers the logic required for the operation, the LLMs can follow this underlying logic to complete the arithmetic operation.}
  \label{fig:Train-data} 
\end{figure}

In our research, the model exhibited exceptional logical reasoning and generalization capabilities while tackling three complex tasks: high-digit(10 or more digits) addition and subtraction, as well as vector cross-product computations. To facilitate these tasks, we created specialized rule datasets for training. For instance, during the training for addition calculations, the model was guided to acquire essential knowledge, including single-digit addition rules, carry rules, digit mapping rules(between numbers and strings), and fundamental computation rules in Fig.~\ref{fig:Train-data}. By mastering these foundational concepts, and following meticulous pre-training \cite{zhang2019ernie}, our model was able to flexibly apply and integrate these rules, enabling it to accurately perform complex mathematical operations, including high-digit addition and subtraction.

 We extended our model's capabilities to perform vector cross-product computations since it had mastered the rules of addition and subtraction. To achieve this goal, we introduced rules for vector representation and cross-product computation into the model to realize vector cross-product calculations. This means that once the model learns these new rules, it could combine the newly acquired rules with existing numerical computation rules to perform vector cross-product calculations. During the process of vector cross-product computation, the model needs to handle a large amount of complex derivation. Through gradual derivation, combining the right-hand rule for cross-products with basic numerical computation rules, the model gains the ability to compute vector cross-products. This strategy showcases the model's deep logical reasoning and strong generalization ability to solve more complex mathematical operations by learning and integrating various rules in TABLE.~\ref{tab:all_rule}. 
 % For the sake of argument, the following are all based on models dealing with ten digits, with the same logic and results for higher digits.

\subsection{Arithmetic Dataset}

We designed the overall computation rule dataset for training, covering a wide range of arithmetic operations from the most basic single-digit arithmetic tasks to various complex arithmetic rules. This dataset, carefully planned, only contains the most basic single-digit operations, including alignment rules, carry rules, borrow rules, basic computation rules, and composite rules. Each arithmetic expression involves $2$ to $10$ operational steps, involving a series of mathematical computation operations, such as addition $(+)$, subtraction $(-)$, and vector cross-product operations $(\times)$. Our constructed dataset contains approximately $20,000$ records.

% \begin{table*}[tp] % 放置表格的位置，h表示here
%     \centering % 居中
%     %\small
%     \caption{Model Parameter Size Comparison Table} % 表格标题
%     %% \caption{模型参数规模对比表} % 表格标题
%     %% \caption{Model Size} % 表格标题
%     \begin{tabular}{ccccccc} % 三列，居中对齐，竖线分隔列
%       \hline % 表格横线
%       MetaRuleGPT Model &  Embedding&Batch size & Heads & Layers & Parameters & Training Steps \\
%       \hline
%       MetaRuleGPT-10M       & \ff{256  }& \ff{10 }& \ff{16   }& \ff{ 5  }& \ff{10}M& \ff{3000} \\
%       MetaRuleGPT-30M      & \ff{256   }& \ff{30 }& \ff{16   }& \ff{ 15 }& \ff{ 30}M& \ff{3000} \\
%       MetaRuleGPT-80M      & \ff{256  }& \ff{50     }& \ff{32   }& \ff{ 20 }& \ff{80}M& \ff{3000}\\
%       MetaRuleGPT-100M      & \ff{256  }& \ff{50     }& \ff{32   }& \ff{ 25 }& \ff{100}M& \ff{3000}\\
  
%       % text-davinci-003 }& \ff{}& -}& -}& -}& -}& - \\
%       \hline
%     \end{tabular}
% \label{tab:config}
% \end{table*}
  
\subsection{MetaRuleGPT Model Structure}
  
\begin{figure}[bp]
  \centering 
  \includegraphics[scale=0.3]{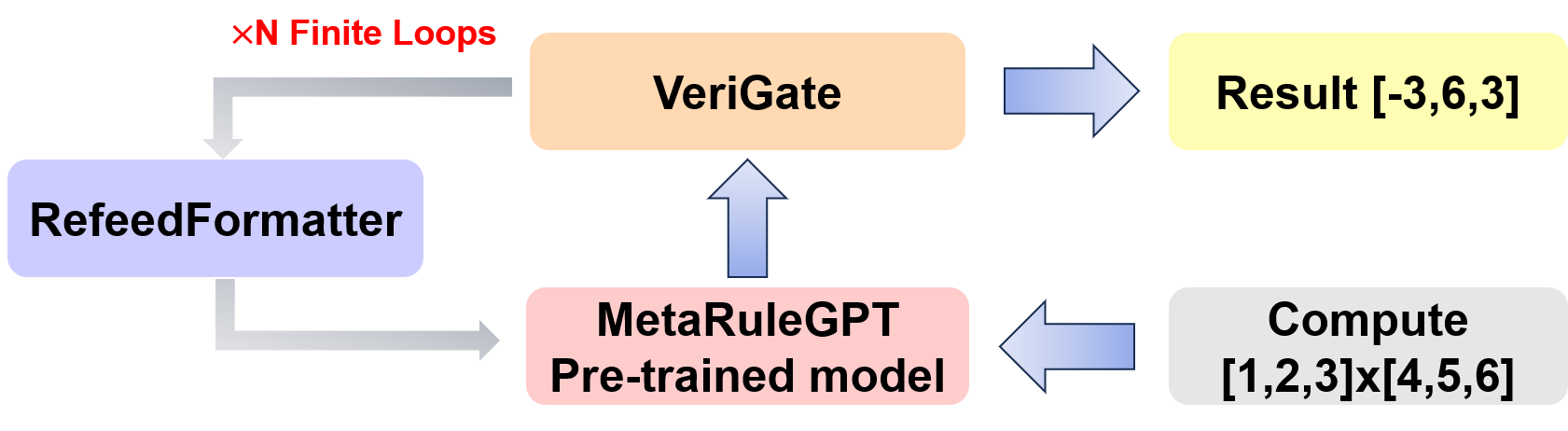}
  \caption{This figure shows Architecture Diagram of MetaRuleGPT. MetaRuleGPT Pre-trained model is a model that has learned basic rules. VeriGate is used to identify whether the current decoding meets the expected transformation of the recent decodings. If it does not meet the expectations, it will enter RefeedFormatter to realign and adjust the structure and then use the basic rules of the model again. According to the operation rules, after a limited number of calls to the basic rule model, the final output is obtained.} 

  \label{fig:model-structure} 
  \end{figure}

To closely mimic the natural process of humans solving mathematical problems, we did not directly solve each complex arithmetic expression but adopted an iterative and stepwise strategy. Through this method, our model breaks down complex expressions into a series of simpler and basic computational steps, reasoning out the final answer step by step. This approach enables the language model to have a deeper understanding and more effective application of specific rules during the learning process, allowing for flexible combination and application of these rules in problem-solving. MetaRuleGPT's proficiency in mathematical tasks stems primarily from its mastery of core computation rules rather than mere memorization of specific cases.
  
Focusing on arithmetic tasks, we developed a language model based on the Transformer architecture, aimed at solving mathematical problems, which we refer to as the MetaRuleGPT language model. The model architecture, as shown in the Fig.~\ref{fig:model-structure}, includes several key components: the MetaRuleGPT pre-trained model, the RefeedFormatter (formatting tool), and VeriGate (verification gate). We designed a self-iteration method that enables the model to simplify complex problems through continuous iteration, and finally obtain the correct answer within a limited number of steps.

\subsection{MetaRuleGPT Pre-trained Model}  
We have trained a language model based on the Transformer\cite{vaswani2017attention} architecture. To flexibly adjust the model's parameter size and internal structure, we designed and implemented a custom Transformer model. Given the limited vocabulary involved in the problems we address, we adopted a single-byte-based training method, which offers clear advantages over traditional word-based or character-based methods.

\begin{figure}[htbp]
    \centering 
    \includegraphics[scale=0.25]{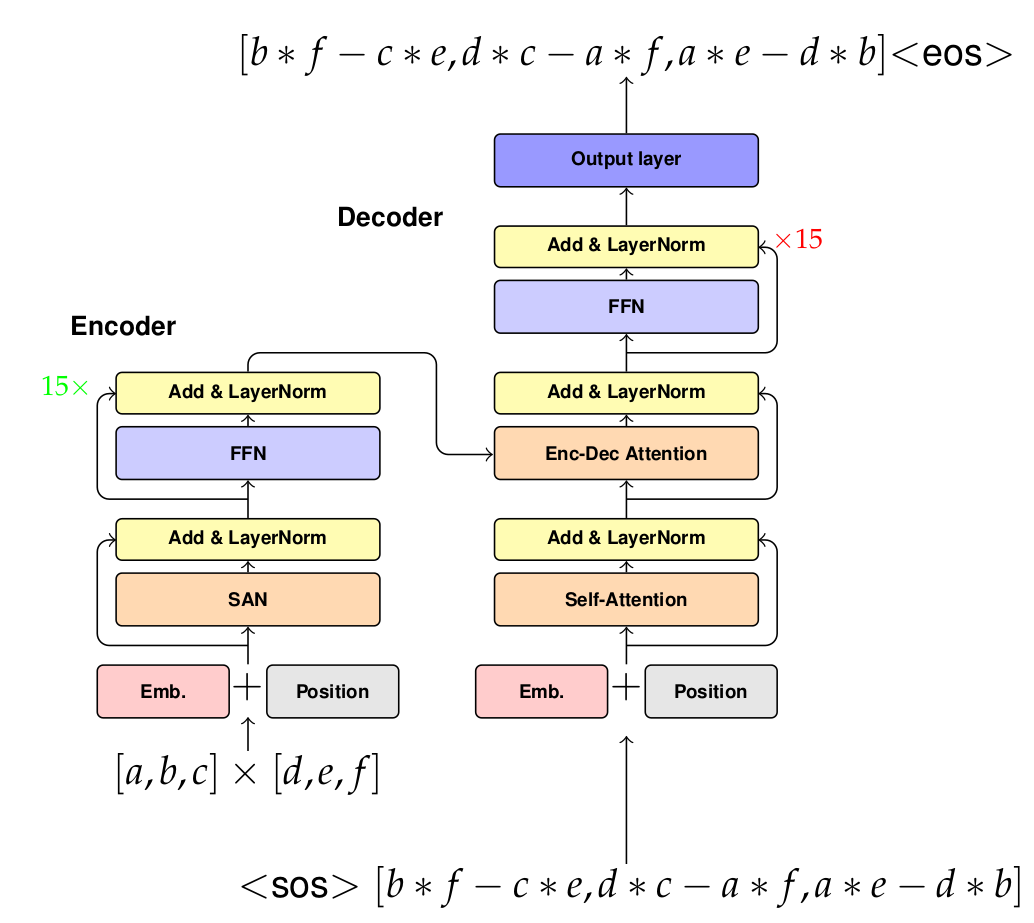}
    \caption{MetaRuleGPT Pre-trained Model. SAN represents Self-Attention Network.} 
    \label{fig:Pre-trained-model} 
    \end{figure}

Byte-based language models provide a flexible and effective means for handling multilingual text and unknown characters. Fig.~\ref{fig:Pre-trained-model} is an example of using the Transformer model to train vector cross product calculation rules. By processing each character individually, the model can ensure more accurate learning of the rules, laying a solid foundation for solving complex logical tasks.
%As shown in Figure \ref{fig:Train-data}, we have trained the dataset using a language model based on the Transformer architecture. To flexibly adjust the model's parameter size and internal structure, we designed and implemented a custom Transformer model. In the design phase of the model, we selected a configuration with 16 attention heads and $15$ layers for both the encoder and decoder, and we specifically chose carefully selected position encoding and word embedding strategies. Given that the problems we face do not involve a complex vocabulary, we adopted a single-byte-based training method. This training strategy has clear advantages and significance compared to traditional word-based or character-based methods.

%Byte-based language models provide a flexible and effective means for handling multilingual text and unknown characters. As shown in Figure \ref{fig:Train-data}, this is an example of using the Transformer model to train vector cross product calculation rules. By processing each character individually, the model can ensure more accurate learning of the rules, laying a solid foundation for solving complex logical tasks.

\subsection{MetaRuleGPT Calculation Example}
\begin{figure*}[htbp]
    \centering % 让图形居中显示
    \includegraphics[width=0.8\textwidth]{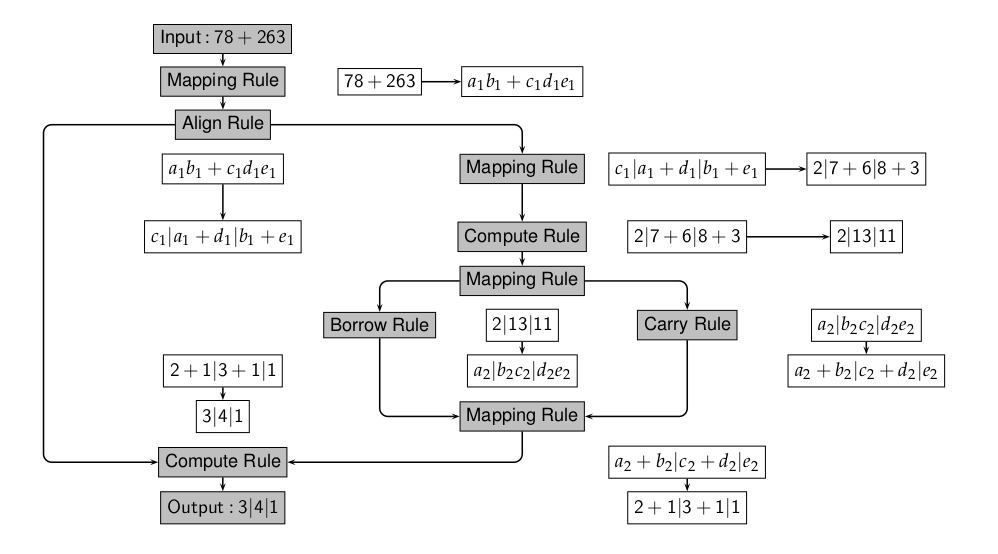}
    \caption{MetaRuleGPT language model calculation example diagram}                    
    \label{fig:example} 
\end{figure*}

To demonstrate how the model operates, we use a simple addition example. When the input ``Input: 78 + 263'' is provided, it is processed sequentially through the Mapping Rule, Compute Rule, Align Rule, and Carry/Borrow Rule to derive the computation result. Fig.~\ref{fig:example} illustrates the internal structure of our model and explains how the initial input is transformed into the final result.

%Figure \ref{fig:Pre-trained-model} illustrates the internal workings of our model. To clearly demonstrate how the model operates, we use a simple addition example. When the input ``Input: 78 + 263'' is provided, it is processed sequentially through the Mapping Rule, Compute Rule, Align Rule, and Carry/Borrow Rule to derive the computation result. Figure \ref{fig:Pre-trained-model} explains how the initial input is transformed into the final result.

\begin{enumerate}
  \item First, the model structurally processes our input question, where ``78 + 263'' under the Mapping Rule becomes:
  
  $$a_1 = 7, b_1 = 8, c_1 = 2, d_1 = 6, e_1 = 3.$$
  
  The expression ``$a_1 b_1 + c_1 d_1 e_1$'' through the Align Rule becomes:
  $$
  c_1 |( a_1 + d_1 )| (b_1 + e_1).
  $$
  Through alignment, a combination of mapping rules produces the intermediate output:``$2|(7 + 6)|(8 + 3)$''.
  
  \item Similarly, for ``$2|(7 + 6)|(8 + 3)$'', a combination of the Mapping Rule and single-digit addition rule (Add Sub-rule) produces the intermediate output: ``$2|13|11$''.
  
  \item When ``$2|13|11$'' is input, the model invokes the Carry Rule and the Mapping Rule to perform digit carry operations, producing an intermediate output: ``$(2 + 1)|(3 + 1)|1$''.
  
  \item ``$(2 + 1)|(3 + 1)|1$'' as a new input, again applying the Mapping Rule and Compute Rule, leads to the final computation result: ``$3|4|1$''.
  
  \item ``$3|4|1$'' as the final input stage, our model invokes the formatting rules and uses special symbols for marking. Finally, the result is formatted using VeriGate to output: ``$Output: 341$''.
  %%Revised the last sentence for better flow
  \end{enumerate}

In summary, MetaRuleGPT utilizes a combination of rules to align, carry, and output the final result during the computation process.

%% file: contents/experiment.tex
\begin{table*}[tp] % 放置表格的位置，h表示here
  \centering % 居中
  %\small
  %% \caption{部分测试数据展示表} % 表格标题
  \caption{Partial Test Data Display Table} % 表格标题 
  \begin{tabular}{c|c c c c c c c} % 三列，居中对齐，竖线分隔列
    \hline % 表格横线
    Data Type & Test Dataset Examples & \\% 表头
    \hline
Randomized Procedure       & \ff{6729132856+1854307391 ,\ldots ,1554887316-817095695 }\\
Perfect Decadic Addition      & \ff{6659891948+340108052,\ldots ,4376628072+623371928}\\
Reverse Magnitude Subtraction  & \ff{62103-2386797965 ,\ldots , 53006-7764286617} \\    
Interleaved Subtraction    &  \ff{1824453209-482835016,\ldots, 8858241744-261714262}\\
Vector Cross Product    &  \ff{(6,5,7)\times(9,3,1),\ldots,(8,2,0)\times(6,4,9)} \\
    \hline
  \end{tabular}
\label{tab:test example}
\end{table*}
\subsection{Experimental Setup}
To demonstrate the exceptional accuracy and generalization capability of our MetaRuleGPT model in reasoning tasks, we designed two experiments: numerical arithmetic tasks and vector cross-product computation tasks. These experiments not only tested the model's basic computational ability but also its ability to solve complex problems, providing a solid foundation for comprehensively evaluating the model's performance in logical reasoning. Furthermore, to further prove the advantages of MetaRuleGPT, we compared it with several well-known large language models, including ChatGPT-3.5, ChatGPT-4.0 \cite{chatgpt2021}, Alibaba's QWen \cite{bai2023qwen}, Google's Palm \cite{anil2023palm,chowdhery2023palm}, Llama2 \cite{touvron2023llama} and Mathematical Mastery Model Goat\cite{liu2023goatfinetunedllamaoutperforms}, demonstrating the superior performance of MetaRuleGPT.

\subsection{Test Dataset}
Current large language models exhibit certain limitations in handling mathematically rigorous problems, partly due to a lack of deep understanding of mathematical logic. In contrast, our model, built from the ground up on the fundamental principles of mathematics, demonstrates higher precision in solving math challenges. To validate this advantage, we designed diverse test datasets and prepared a comprehensive computation dataset to highlight the significant advantage of our model in mathematical reasoning.
Through these validations, we demonstrate that MetaRuleGPT not only precisely grasps and applies the basic logic of mathematics but also exhibits powerful generalization in problem-solving.

In the domain of arithmetic tasks, we constructed a diverse training dataset containing a wide range of arithmetic operations. To comprehensively evaluate our model's computational accuracy and generalization ability, we designed an evaluation dataset containing $8,000$ test cases shown in Table.~\ref{tab:test example}, entirely non-overlapping with the training set. This dataset covers various types of numerical operations, including but not limited to perfect decimal addition, reverse magnitude subtraction, misplaced subtraction, and addition and subtraction operations based on randomly generated numbers.

To compare with other models focusing on mathematical calculations, we used an additional substantial number of randomly generated dataset sampled from a logarithmic space\footnote{Our dataset is available at \url{https://www.scidb.cn/en/detail?dataSetId=04575028fb8d4bfabeeba5825c8f57fc}}. This sampling approach ensures that the numbers are equally likely to originate from different orders of magnitude, similar to the method employed by \cite{lee2023recursion}.

\begin{figure}[tp]
  \centering
  \includegraphics[scale=0.4]{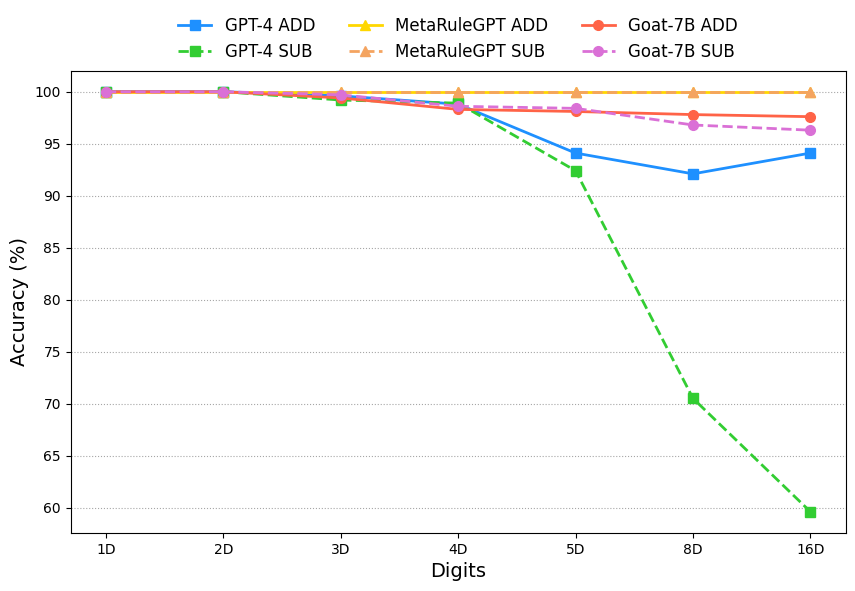}
  \caption{Comparison of the accuracy of addition and subtraction of GPT4, Goat and MetaRuleGPT on different digits. When the number of digits increases, the accuracy of other models begins to decline, indicating that they do not understand the principles of arithmetic. Our model can guarantee 100\% accuracy.}
  \label{fig:goat} 
\end{figure}

\begin{table}[tp] % 放置表格的位置，h表示here
  \centering % 居中
  %% \caption{Interleaved Subtraction(交错式位值减法)} % 表格标题
\caption{Language Models' Overall Performance in Numerical Tasks} 
  \begin{tabular}{cccccc} % 三列，居中对齐，竖线分隔列
    \hline % 表格横线
    Model & Model Parameter & 5-digit & 10-digit   \\
    \hline
    GPT-4& \ff{1760}B$+$  & \ff{99.22}\%     & \ff{90.9}\%  \\
    GPT-3.5& \ff{175}B$+$    & \ff{97.26}\%    & \ff{83.9}\%  \\
    Llama2-7b& \ff{7}B   & \ff{22.3}\%      & \ff{1.7}\%   \\    
    Llama2-13b& \ff{13}B  & \ff{17.8}\%    & \ff{ 1.6}\%   \\    
    Llama2-70b& \ff{70}B  & \ff{57.76}\%    & \ff{ 6.4}\%  \\
    Google-PaLM& \ff{110}B & \ff{73.32}\%      & \ff{26.6}\%  \\
    Qwen-72b-Chat& \ff{72}B & \ff{91.32}\%    & \ff{60.4}\%   \\
    \hline
    \hline
    MetaRuleGPT & \ff{30}M & {$100$}\% & {\ff{100}}\%  \\ 
    \hline
  \end{tabular}
  \label{tab:mean}
\end{table}

\subsection{Evaluation Metrics}
In evaluating the model's performance, we consider not only the accuracy of the computed results but also the difference ratio between the computed and correct answers. A smaller absolute difference ratio indicates that the model's output is closer to the true value, suggesting potential for improvement through parameter tuning. Conversely, a difference ratio greater than 1 implies that the model struggles with such problems or faces significant challenges.
Assuming the number of correctly predicted quantities is TP and the total number of predictions is N, then accuracy can be defined as:
{Accuracy} $= \frac{TP}{N} \times 100\%$

Suppose our model's computation result is y, the actual computation result N numbers in total, then our final overall difference ratio can be defined as:\\
{DifferenceRatio} $= \frac{1}{N} \sum_{i=0}^N \left| \frac{y_i - \hat{y_i}}{\max(y_i, \hat{y_i})} \right|$

%Assuming the number of correctly predicted quantities is TP and the total number of predictions is N, then accuracy can be defined as:${Accuracy} = \frac{TP}{N} \times 100\%$

%Suppose our model's computation result is y, the actual computation result N numbers in total, then our final overall difference ratio can be defined as:
%${Difference Ratio} = \frac{1}{N} \sum_{i=0}^N \left| \frac{y_i - \hat{y_i}}{\max(y_i, \hat{y_i})} \right|$
\begin{table}[tp] % 放置表格的位置，h表示here
  \centering % 居中
  \small
  \caption{Interleaved Subtraction} % 表格标题
  %% \caption{交错式位值减法表} % 表格标题
  \begin{tabular}{ccccccc} % 三列，居中对齐，竖线分隔列
    \hline % 表格横线
    Compute Digits & \multicolumn{2}{c}{5-digit} &\multicolumn{2}{c}{10-digit}  \\
    \hline % 表格横线
    Rate & Error & Accuracy & Error & Accuracy  \\
    \hline
    GPT-4         & \ff{0.016} & \ff{98.3}\%     & 5.368e-7 & \ff{ 96}\% \\
    GPT-3.5       & \ff{0.0033} & \ff{95.2}\%    & \ff{0.037} & \ff{   91}\%   \\
    Llama2-7b     & \ff{0.64} & \ff{2.3}\%      & \ff{0.92} & \ff{  0}\%    \\    
    Llama2-13b    & \ff{0.52} & \ff{21.9}\%    & \ff{0.69} & \ff{   2}\%   \\    
    Llama2-70b    & \ff{0.061} & \ff{76.1}\%    & \ff{0.79} & \ff{   2}\%      \\
    Google-PaLM   & \ff{0.0076} & \ff{95.9}\%      & \ff{0.54} & \ff{19}\%  \\
    Qwen-72b-Chat & \ff{0.0092} & \ff{93.9}\%    & \ff{0.0027} & \ff{74.5}\%    \\
    \hline
    \hline
    MetaRuleGPT & \textbf{\ff{0.0}} & \textbf{\ff{100}}\% & \textbf{\ff{0.0}} & \textbf{\ff{100}}\% \\
    \hline
  \end{tabular}
  \label{tab:interleaved}
\end{table}

\begin{table}[tp] % 放置表格的位置，h表示here
  \centering % 居中
  \small
  \caption{Reverse Magnitude Subtraction} % 表格标题
    %% \caption{反向幅度减法表} % 表格标题 
  \begin{tabular}{ccccccc} % 三列，居中对齐，竖线分隔列
    \hline % 表格横线
    Compute Digits & \multicolumn{2}{c}{5-digit} &\multicolumn{2}{c}{10-digit}  \\
    \hline % 表格横线
    Rate & Error & Accuracy & Error & Accuracy  \\
    \hline
    GPT-4         & \ff{0.027} & \ff{97.8}\%          & 1.3e-8 & \ff{ 96.5}\% \\
    GPT-3.5       & \ff{0.0033} & \ff{99.4}\%    & 8.3e-4 & \ff{  88.5}\% \\
    Llama2-7b     & \ff{0.64} & \ff{20.8}\%      & \ff{2.1} & \ff{ 0.0}\%        \\    
    Llama2-13b    & \ff{0.52} & \ff{4}\%    & \ff{1.5} & \ff{    0.0}\%   \\    
    Llama2-70b    & \ff{0.061} & \ff{50.2}\%    & 4.6e-6 & \ff{   0.5}\%      \\
    Google-PaLM   & \ff{0.0076} & \ff{43.6}\%       & \ff{1.0} & \ff{0.0}\%  \\
    Qwen-72b-Chat & \ff{0.0092} & \ff{86.4}\%    & \ff{0.065} & \ff{3.5}\%    \\
    \hline
    \hline
    MetaRuleGPT & \textbf{\ff{0.0}} & \textbf{\ff{100}}\% & \textbf{\ff{0.0}} & \textbf{\ff{100}}\% \\
    \hline
  \end{tabular}
  \label{tab:reverse}
\end{table}

\subsection{Deep Numerical Optimization Experiments on Language Models}
To test our model's mathematical reasoning and generalization capabilities, we conducted comparisons using well-known language models such as the currently leading ChatGPT-4.0, ChatGPT-3.5, Alibaba's QWen, Google's Palm, Llama2 and Goat. Through such comparisons, we can comprehensively assess the performance differences between models and evaluate MetaRuleGPT's capabilities in mathematical reasoning tasks. A comprehensive comparison is shown in Fig.~\ref{fig:goat} and Table.~\ref{tab:mean}.

Using the specific test datasets, we previously organized to invoke and test with the aforementioned large language models, preserving and comparing the computational results of each model. We conducted a series of detailed experiments and evaluations, and the results are shown in Table.~\ref{tab:interleaved}, \ref{tab:reverse}, \ref{tab:random-sum}, \ref{tab:perfect-addition} and \ref{tab:random-sub}.

Finally, in order to compare with the public dataset, we selected some subsets of gsm8k\cite{cobbe2021gsm8k} and simplified the natural language part into mathematical formulas for comparison. The results are shown in \ref{tab:gsm8k}.

\subsection{Language Model-Driven Vector Cross Product Calculation Experiment}
To demonstrate our model's capability in handling complex logical problems, we have selected vector cross product calculation, a challenging mathematical task, as a test case. This test not only verifies the model's computational accuracy but also compares its performance with those of leading large language models. Table.~\ref{tab:vector-cross} presents the detailed accuracy comparison results of various models on the vector cross product calculation dataset.

%% file: contents/Results_and_Discussion.tex
\subsection{Test Data Results Analysis}

\begin{table}[tp] % 放置表格的位置，h表示here
  \centering % 居中
  \small
  \caption{Randomized Addition Procedure} % 表格标题
  %% \caption{对比实验-随机加法表} % 表格标题
  \begin{tabular}{ccccccc} % 三列，居中对齐，竖线分隔列
    \hline % 表格横线
    Compute Digits & \multicolumn{2}{c}{5-digit} &\multicolumn{2}{c}{10-digit}  \\
    \hline % 表格横线
    Rate & Error & Accuracy & Error & Accuracy  \\
    \hline
    GPT-4         & \textbf{\ff{0.0}} & \ff{100}\%   & \ff{0.092} & \ff{ 85.5}\% \\
    GPT-3.5       & 1.6e-5 & \ff{99}\%    & \ff{0.046} & \ff{ 72}\%   \\
    Llama2-7b     & \ff{0.7434} & \ff{49.5}\%   & \ff{6.6} & \ff{  0.5}\%  \\    
    Llama2-13b    & \ff{0.5075} & \ff{28.5}\%    & \ff{0.47} & \ff{  2}\%       \\
    Llama2-70b    & \ff{0.0165} & \ff{84}\%    & \ff{0.64} & \ff{   11}\%      \\
    Google-PaLM   & \ff{0.0017} & \ff{94}\%         & \ff{0.34} & \ff{39.5}\%  \\
    Qwen-72b-Chat & \ff{0.0005} & \ff{97}\%    & \ff{0.0082} & \ff{75}\%    \\
    \hline
    \hline
    MetaRuleGPT & \textbf{\ff{0.0}} & \textbf{\ff{100}}\% & \textbf{\ff{0.0}} & \textbf{\ff{100}}\% \\
    \hline
  \end{tabular}
  \label{tab:random-sum}
\end{table}

\begin{table}[tp] % 放置表格的位置，h表示here
  \centering % 居中
  \small
  \caption{Randomized Subtraction Procedure} % 表格标题
  %% \caption{对比实验-随机减法表} % 表格标题
  \begin{tabular}{ccccccc} % 三列，居中对齐，竖线分隔列
    \hline % 表格横线
    Compute Digits & \multicolumn{2}{c}{5-digit} &\multicolumn{2}{c}{10-digit}  \\
    \hline % 表格横线
    Rate & Error & Accuracy & Error & Accuracy  \\
    \hline
    GPT-4         & \textbf{\ff{0.0}} &\ff{ 100}\%     & \ff{0.019} &\ff{  78.5}\% \\
    GPT-3.5       & 4.9e-5 &\ff{ 95.5}\%    & \ff{0.052} &\ff{  76.5}\% \\
    Llama2-7b     & \ff{1.4} &\ff{ 25.5}\%    & \ff{2.1} &\ff{  3}\%     \\    
    Llama2-13b    & \ff{0.49} &\ff{ 31}\%      & \ff{0.67} &\ff{  1}\%    \\    
    Llama2-70b    & \ff{0.085} &\ff{ 73.5}\%    & \ff{1.5} &\ff{  12.5}\%    \\
    Google-PaLM   & \ff{0.15} &\ff{ 80.5}\%    & \ff{0.68} &\ff{  47}\%  \\
    Qwen-72b-Chat & 2.0e-4 &\ff{ 93.5}\%    & \textbf{\ff{0.0017}} &\ff{  81.5}\%    \\
    \hline
    \hline    
    MetaRuleGPT  & \textbf{\ff{0.0}} &\textbf{\ff{ 100}}\%     & \ff{0.063} &\textbf{\ff{100}}\%    \\
    \hline
  \end{tabular}
  \label{tab:random-sub}
\end{table}

\begin{table}[bp] % 放置表格的位置，h表示here
  \centering % 居中
  %% \caption{Vector Cross Product(向量外积计算)} % 表格标题
  \caption{Simplified gsm8k Table} % 表格标题 
  \begin{tabular}{lccc} % 三列，居中对齐，竖线分隔列
    \hline % 表格横线
    Model & Accuracy       \\
    \hline
    GPT-4         & \ff{100}\%  \\
    GPT-3.5       & \ff{99}\%  \\
    llama2-7b     & -  \\
    llama2-13b    & -   \\
    llama2-70b    & \ff{98}\%    \\
    Google-PaLM   & \ff{100}\% \\
    Qwen-72b-Chat & \ff{100}\%   \\
    \hline
    \hline
    MetaRuleGPT  & \textbf{\ff{100}}\%    \\
    \hline
  \end{tabular}
\label{tab:gsm8k}
\end{table}

\subsubsection{Numerical calculation results}
To assess our model's performance in solving general numerical problems, we generated a large amount of experimental data with random numbers using Python. The preliminary results show that MetaRuleGPT and other tested language models achieved accuracy rates exceeding $70\%$ in low-digit addition and subtraction operations, demonstrating high computational precision. However, as the number of digits increased, the performance of most language models significantly declined. Except for ChatGPT, other models often made mistakes in handling high-digit calculations due to their inability to deeply grasp computational rules, almost losing their computational capability.

Remarkably, MetaRuleGPT maintained $100\%$ accuracy even when facing high-digit random addition tasks. Although it faced certain challenges in high-digit random subtraction tasks, our model still showed the highest accuracy among all tested language models. This achievement not only highlights MetaRuleGPT's strong performance in solving complex numerical problems but also proves its generalization ability.

Table \ref{tab:interleaved} presents the test results of various models on misaligned subtraction. 
Table \ref{tab:reverse} shows the test results of various models on reverse amplitude subtraction.
Tables \ref{tab:random-sum} and \ref{tab:random-sub} respectively demonstrate the test results of various models on randomly generated numbers.
Table \ref{tab:perfect-addition} displays the test results of various models on perfect decimal addition.

\subsubsection{Vector Cross Product Results}
From the data in Table \ref{tab:mean}, it is evident that the Llama2 models with 7B and 13B parameter sizes were incapable of performing vector cross product calculations, while the Llama2-70B, the largest parameter model of the Llama2 series, could perform cross product calculations but with $0\%$ accuracy. Even the state-of-the-art ChatGPT achieved an accuracy rate below $50\%$ without the aid of external tools. In contrast, MetaRuleGPT accurately calculated vector cross products in three-dimensional space with $100\%$ accuracy, confirming the effectiveness of enhancing model capabilities by combining different rules. By comprehensively learning basic operational rules such as addition, subtraction, multiplication, and cross product, MetaRuleGPT achieved precise invocation of these rules and successfully outputted accurate calculation results.
Table \ref{tab:vector-cross} provides detailed accuracy comparison results of different models on the dataset for vector cross-product computation.

\begin{table}[tp] % 放置表格的位置，h表示here
  \centering % 居中
  \small
  \caption{Perfect Decadic Addition} % 表格标题
  %% 
  %% \caption{完美十进制加法表} % 表格标题
  \begin{tabular}{ccccccc} % 三列，居中对齐，竖线分隔列
    \hline % 表格横线
    Compute Digits & \multicolumn{2}{c}{5-digit} &\multicolumn{2}{c}{10-digit}  \\
    \hline % 表格横线
    Rate & Error & Accuracy & Error & Accuracy  \\
    \hline 
    GPT-4         & \textbf{\ff{0.0}} & \ff{ 100}\%          & 2.9e-9 & \ff{ 98}\%  \\
    GPT-3.5       & 4.2e-5 & \ff{97.2}\%    & 2.0e-4 & \ff{   91.5}\%  \\
    Llama2-7b     & \ff{15} & \ff{13.4}\%      & \ff{30} & \ff{ 5}\%     \\
    Llama2-13b    & \ff{0.16} & \ff{3.6}\%    & \ff{10} & \ff{  3}\%    \\    
    Llama2-70b    & \ff{0.14} & \ff{5}\%    & \ff{1.8} & \ff{ 6}\%    \\
    Google-PaLM   & \ff{0.89} & \ff{ 52.6}\%         & \ff{24} & \ff{27.5}\%  \\
    Qwen-72b-Chat & \ff{0.27} & \ff{85.8}\%    & \ff{0.21} & \ff{67.5}\%    \\
    \hline
    \hline
    MetaRuleGPT  & \textbf{\ff{0.0}} & \textbf{\ff{100}}\%     & \textbf{\ff{0.0}} & \textbf{\ff{100}}\%    \\
    \hline
  \end{tabular}
\label{tab:perfect-addition}
\end{table}

\begin{table}[bp] % 放置表格的位置，h表示here
  \centering % 居中
  %% \caption{Vector Cross Product(向量外积计算)} % 表格标题
  \caption{Vector Cross Product Table} % 表格标题 
  \begin{tabular}{lccc} % 三列，居中对齐，竖线分隔列
    \hline % 表格横线
    Vector Compute & Accuracy        \\
    \hline
    GPT-4         & \ff{17}\%  \\
    GPT-3.5       & \ff{5.5}\%  \\
    llama2-7b     & -  \\
    llama2-13b    & -   \\
    llama2-70b    & \ff{0}\%    \\
    Google-PaLM   & \ff{0}\% \\
    Qwen-72b-Chat & \ff{23}\%   \\
    \hline
    \hline
    MetaRuleGPT  & \textbf{\ff{100}}\%    \\
    \hline
  \end{tabular}
\label{tab:vector-cross}
\end{table}

%The specific experimental results can be found in the appendix A.1.

%To assess our model's performance in solving general numerical problems, we generated a large amount of experimental data with random numbers using Python. Preliminary results show that in low-digit addition and subtraction operations, our model and other tested language models achieved an accuracy rate exceeding 70\%, demonstrating high computational precision. However, as the number of digits increased, the performance of most language models significantly declined. Except for ChatGPT, other models often made mistakes in handling high-digit calculations due to their inability to deeply grasp computational rules, nearly losing their computational capability.

%{It is particularly worth mentioning that even when facing high-digit random addition tasks, our model still maintained a 100\% accuracy rate. Although it faced certain challenges in high-digit random subtraction tasks, our model still showed the highest accuracy among all tested language models. This achievement not only highlights our model's good performance in solving complex numerical problems but also proves its generalization ability.}

% \hl{[The claims about 100\% accuracy and 10\% higher accuracy than ChatGPT may raise questions from reviewers, as they seem quite high. It would be better to provide more specific evidence or data to support these claims, or tone them down slightly to avoid potential skepticism.]}

%The specific experimental results can be found in the appendix A.1.

% \input{tables/cross_prod.tex}

Moreover, by training rules for two different types of tasks within the same pre-trained model, MetaRuleGPT demonstrated multi-task generalization ability. This indicates that our model is not only adaptable to various task scenarios but can also identify and apply common rules among these tasks, significantly enhancing learning efficiency. These results further showcase MetaRuleGPT's strong performance and flexibility.

%From the data in Table \ref{tab:mean}, it is evident that the Llama2 models with 7b and 13b parameter sizes were even unable to perform vector cross product calculations, while the Llama2-70b, the largest parameter model of the Llama2 series, could perform cross product calculations but with an accuracy rate of 0\%. Even the currently most powerful language model, ChatGPT, achieved an accuracy rate below 50\% without the aid of external tools. In contrast, our model was able to accurately calculate vector cross products in three-dimensional space with an accuracy rate of 98.5\%, further confirming that the method of enhancing model capabilities by combining different rules is effective. By comprehensively learning basic operational rules such as addition, subtraction, multiplication, and cross product, our model achieved precise invocation of these rules and successfully outputted accurate calculation results. More importantly, by training rules for two different types of tasks within the same pre-trained model, our model demonstrated multi-task generalization ability. This indicates that our model is not only adaptable to a variety of different task scenarios but can also identify and apply common rules among these tasks, significantly enhancing learning efficiency. This further showcases our model's good performance and flexibility.

\subsection{Discussion}

%\subsubsection{Controllability}
Although existing language models have demonstrated powerful capabilities, they still face challenges in terms of controllability. In particular, most models struggle to precisely answer questions within a controlled range, often resulting in significant deviations, which is a crucial issue that current language models need to address. Conversely, MetaRuleGPT's rule-based execution ensures relative reliability and better controllability. 

Experiments show that existing language models struggle with high-digit calculations and complex computational tasks due to limitations in understanding and the difficulty of learning a unified representation for text and numbers. MetaRuleGPT addresses these challenges by precisely completing complex tasks and demonstrating the generalization potential of language models in numerical calculations through rule learning.